\documentclass[sigconf,natbib=true]{acmart}

\setcctype{}

\setcopyright{none}
\renewcommand\footnotetextcopyrightpermission[1]{}

\usepackage{algorithm}
\usepackage{multirow}
\usepackage{multicol}
\usepackage{amsmath, bm}
\usepackage{mathtools, physics}
\usepackage{color}
\usepackage{graphicx}   
\usepackage{siunitx}    
\usepackage[table]{xcolor} 
\usepackage{xcolor}
\usepackage{colortbl}
\usepackage{algpseudocode}
\usepackage{tabularx}
\usepackage{booktabs}
\usepackage{array}
\usepackage{enumitem}
\usepackage{xcolor}
\usepackage{pifont}
\usepackage{ragged2e}
\usepackage{threeparttable}

\usepackage[most]{tcolorbox}   
\usepackage{caption}           

\newtcolorbox{promptbox}[1][]{
  enhanced, breakable,
  colback=white,            
  colframe=black!60,        
  arc=2mm, boxrule=0.6pt,   
  drop shadow,              
  title=Prompt for Query Classification,
  colbacktitle=gray!30,     
  coltitle=white,           
  fonttitle=\bfseries\large,
  toptitle=2mm, bottomtitle=1mm,
  left=3mm, right=3mm, top=2mm, bottom=2mm,
  #1
}

\newcolumntype{Y}{>{\centering\arraybackslash}X} 
\newcolumntype{L}{>{\raggedright\arraybackslash}X} 
\newcolumntype{C}{>{\centering\arraybackslash}X}   
\newcolumntype{Z}{>{\ttfamily\footnotesize\RaggedRight\arraybackslash}X} 
\newcolumntype{W}[1]{>{\hsize=#1\hsize\RaggedRight\arraybackslash}X} 
\newcolumntype{P}[1]{>{\RaggedRight\arraybackslash}p{#1}}
\newcolumntype{T}[1]{>{\ttfamily\footnotesize\RaggedRight\arraybackslash}p{#1}}

\newcommand{\cmark}{{\ding{51}}}
\newcommand{\xmark}{{\ding{55}}}
\renewcommand\tabularxcolumn[1]{m{#1}} 

\AtBeginDocument{%
  }

\setcopyright{acmlicensed}
\copyrightyear{2018}
\acmYear{2018}
\acmDOI{XXXXXXX.XXXXXXX}
\acmConference[Conference]{}{Under Review}{}
\acmISBN{978-1-4503-XXXX-X/2018/06}

\begin{document}

\title{Use Graph When It Needs: Efficiently and Adaptively Integrating Retrieval-Augmented Generation with Graphs}

\author{Su Dong}
\email{su.dong@connect.polyu.hk}
\affiliation{%
  \institution{The Hong Kong Polytechnic
University}
  \city{Hung Hom}
  \state{Kowloon}
  \country{Hong Kong}
}

\author{Qinggang Zhang}
\email{qinggang.zhang@polyu.edu.hk}
\affiliation{%
  \institution{The Hong Kong Polytechnic
University}
  \city{Hung Hom}
  \state{Kowloon}
  \country{Hong Kong}
}

\author{Yilin Xiao}
\affiliation{%
  \institution{The Hong Kong Polytechnic
University}
  \city{Hung Hom}
  \state{Kowloon}
  \country{Hong Kong}
}

\author{Shengyuan Chen}
\affiliation{%
  \institution{The Hong Kong Polytechnic
University}
  \city{Hung Hom}
  \state{Kowloon}
  \country{Hong Kong}
}

\author{Chuang Zhou}
\affiliation{%
  \institution{The Hong Kong Polytechnic
University}
  \city{Hung Hom}
  \state{Kowloon}
  \country{Hong Kong}
}

\author{Xiao Huang}
\email{xiao.huang@polyu.edu.hk}
\affiliation{%
  \institution{The Hong Kong Polytechnic
University}
  \city{Hung Hom}
  \state{Kowloon}
  \country{Hong Kong}
}

\renewcommand{\shortauthors}{Dong et al.}



\begin{abstract}
Large language models (LLMs) often struggle with knowledge-intensive tasks due to hallucinations and outdated parametric knowledge. While Retrieval-Augmented Generation (RAG) addresses this by integrating external corpora, its effectiveness is limited by fragmented information in unstructured domain documents. Graph-augmented RAG (GraphRAG) emerged to enhance contextual reasoning through structured knowledge graphs, yet paradoxically underperforms vanilla RAG in real-world scenarios, exhibiting significant accuracy drops and prohibitive latency despite gains on complex queries. We identify the rigid application of GraphRAG to all queries, regardless of complexity, as the root cause. To resolve this, we propose an efficient and adaptive GraphRAG framework called EA-GraphRAG that dynamically integrates RAG and GraphRAG paradigms through syntax-aware complexity analysis. Our approach introduces: (i) a syntactic feature constructor that parses each query and extracts a set of structural features; (ii) a lightweight complexity scorer that maps these features to a continuous complexity score; and (iii) a score-driven routing policy that selects dense RAG for low-score queries, invokes graph-based retrieval for high-score queries, and applies complexity-aware reciprocal rank fusion to handle borderline cases. Extensive experiments on a comprehensive benchmark, consisting of two single-hop and two multi-hop QA benchmarks, demonstrate that our EA-GraphRAG significantly improves accuracy, reduces latency, and achieves state-of-the-art performance in handling mixed scenarios involving both simple and complex queries. 

\end{abstract}

\maketitle

\section{Introduction}
Large Language Models (LLMs) have significantly advanced the field of natural language processing, yet their tendency to generate inaccurate or outdated information remains a critical challenge. To mitigate this, Retrieval-Augmented Generation (RAG)~\cite{gao2023retrieval,lewis2020retrieval} has emerged as a prominent technique, enriching LLMs with real-time access to external knowledge sources. By retrieving relevant text passages from the external corpus using semantic matching, RAG grounds model responses in verifiable evidence, significantly improving accuracy for knowledge-intensive tasks.

\begin{figure}[t] 
    \centering
    \includegraphics[width=0.49\textwidth]{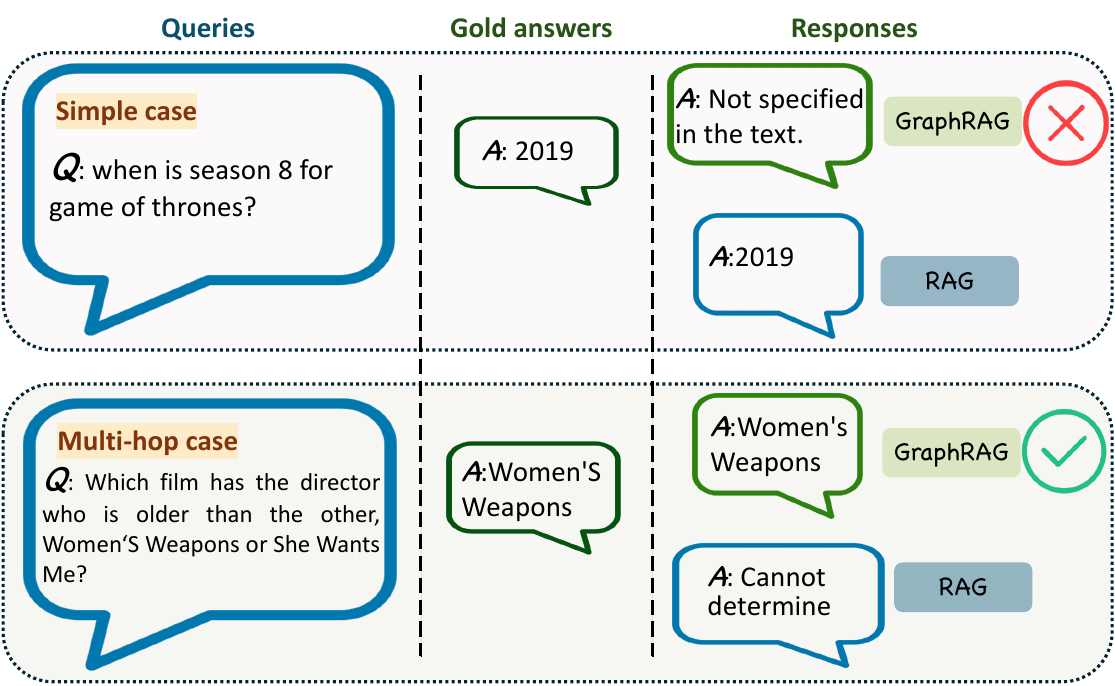}
    \caption{An illustrative example showing that GraphRAG often underperforms vanilla RAG on simple single-hop queries, but excels on multi-hop queries requiring relational reasoning; conversely, vanilla RAG tends to perform better on single-hop queries while struggling with multi-hop reasoning.}
    \label{fig:example}
\end{figure}

Despite the success, traditional RAG systems often fall short when faced with complex tasks that require reasoning across multiple pieces of interconnected information, such as those involving causal relationships or multi-step logical inferences. To this end, graph retrieval-augmented generation (GraphRAG)~\cite{zhang2025survey,procko2024graph,peng2024graph} has recently emerged as a powerful paradigm that leverages external structured graphs to enable deep retrieval and contextual comprehension. GraphRAG can effectively answer questions about indirect influences or causal chains by exploring connections between nodes. Early GraphRAG approaches established the core methodology. Microsoft GraphRAG~\cite{edge2024local} pioneered the use of a globally constructed knowledge graph, employing community detection algorithms to identify thematic clusters and enabling comprehensive querying through a combination of local (entity-specific) and global (community-level) searches. Building on this foundation, RAPTOR~\cite{sarthi2024raptor} introduced a recursive, tree-based indexing strategy as an alternative to graphs, creating a hierarchy of summarized text chunks to facilitate multi-scale reasoning without relying on explicit entity-relationship extraction. Subsequent research focused on enhancing the efficiency and biological plausibility of retrieval. HippoRAG~\cite{HippoRAG} and its successor HippoRAG2~\cite{gutiérrez2025hipporag2} marked a significant departure by modeling human memory processes, using a neurobiologically-inspired framework that integrates an LLM with a knowledge graph and the Personalized PageRank algorithm for efficient, single-step associative recall. Parallel efforts aimed at optimization led to LightRAG~\cite{guo2024lightrag}, which maintains a graph's relational benefits while drastically improving scalability through a dual-level retrieval system and a more lightweight, graph-enhanced indexing process. Collectively, these strategies in the GraphRAG lineage have significantly advanced retrieval precision and contextual depth, empowering LLMs to tackle intricate, multi-hop queries more effectively than ever before.


However, GraphRAG models still suffer from several key issues in real-world scenarios. (i) Low efficiency: Most existing GraphRAG models rely heavily on LLM-based graph construction, which incurs significant token costs and update latency, making them impractical for large-scale or dynamically evolving knowledge bases. (ii) Poor generalizability: GraphRAG exhibits poor generalizability on simpler tasks. While GraphRAG outperforms RAG models on complex reasoning tasks, they frequently underperform traditional RAG approaches on many simple NLP tasks~\cite{han2025rag,zhou2025depth}. Empirical evidence~\cite{xiang2025use,zhou2025depth} shows that GraphRAG underperforms vanilla RAG on single-hop question-answering, achieving 13.4\% lower accuracy on Natural Questions and a 16.6\% drop for time-sensitive queries. It is because while graph-based retrieval improves reasoning depth, it also introduces noise and ambiguities into the retrieved contexts.



These observations highlight a fundamental trade-off: while GraphRAG excels in complex reasoning, its inefficiencies and limitations in handling simple queries make it suboptimal as a universal solution. An ideal system should dynamically route queries to the appropriate paradigm based on reasoning complexity, i.e., employing RAG for simplicity and efficiency, and GraphRAG for depth and complexity. However, it is hard to (i) accurately and efficiently quantify query complexity, and (ii) build a lightweight mechanism to perform this routing.

To bridge this gap, we introduce EA-GraphRAG, a novel framework that adaptively integrates RAG with graphs by effectively evaluating query syntax complexity. Our approach consists of three key components: (i) a syntactic feature constructor that parses each query and extracts structural features, (ii) a lightweight complexity scorer that maps these features to a continuous complexity score, and (iii) a score-driven routing policy that selects dense RAG for low-complexity queries, invokes GraphRAG for high-complexity queries, and applies complexity-aware fusion for borderline cases. This syntax-aware strategy maximizes efficiency while preserving reasoning depth. Extensive experiments on two single-hop and two multi-hop QA benchmarks verify EA-GraphRAG’s effectiveness in balancing accuracy, latency, and contextual fidelity. To summarize, our contribution is listed as follows:
\begin{itemize}
    \item We identify effectiveness and efficiency limitations in existing GraphRAG methods for simple single-hop QA tasks. Based on these findings, we propose EA-GraphRAG, a novel framework that dynamically integrates RAG with graphs by evaluating query syntactic complexity.

    \item EA-GraphRAG introduces a lightweight query complexity analyzer to evaluate the complexity of user queries.
    
    \item Based on the syntax complexity, EA-GraphRAG designs an adapter that routes simple queries to vanilla RAG while reserving GraphRAG for complex reasoning.
    
    \item Extensive experiments on comprehensive benchmarks consisting of both single-hop QA datasets and multi-hop QA datasets demonstrate that EA-GraphRAG outperforms the state-of-the-art baselines.
\end{itemize}

\section{Related Work}
\subsection{Retrieval-Augmented Generation}
Retrieval-Augmented Generation (RAG) has emerged as a key paradigm to address the limitations of large language models (LLMs) in knowledge-intensive tasks, enabling access to external corpora to enhance response accuracy and reduce hallucinations ~\cite{gao2023retrieval,lewis2020retrieval}. Early advancements in RAG focused on optimizing retrieval mechanisms: BM25 ~\cite{robertson1994some} leverages term frequency statistics for efficient text matching, while dense retrievers like Contriever ~\cite{izacard2021unsupervised} and ColBERTv2 ~\cite{santhanam2022colbertv2} use contextual embeddings to improve semantic alignment. More recent methods, such as IRCoT ~\cite{trivedi2023interleaving}, integrate retrieval with chain-of-thought reasoning to handle multi-step queries, and Atlas~\cite{izacard2023atlas} enhances few-shot learning by aligning retrieval with in-context examples. However, traditional RAG systems struggle with large-scale, unstructured domain corpora, where information fragmentation and loss of contextual relationships during chunking compromise performance on complex reasoning tasks ~\cite{han2024retrieval,zhang2025survey}.
\subsection{Graph Retrieval-Augmented Generation}
To mitigate RAG’s limitations in contextual comprehension, Graph-based RAG integrates structured knowledge graphs to model conceptual relationships~\cite{peng2024graph,procko2024graph}, enhancing reasoning depth. Key innovations include hierarchical community-based search in Microsoft GraphRAG ~\cite{edge2024local} and its variant LazyGraphRAG~\cite{LazyGraphRAG}, which combine local and global querying for comprehensive responses. LightRAG~\cite{guo2024lightrag} improves scalability through dual-level retrieval and graph-enhanced indexing, while GRAG~\cite{hu2024grag} introduces soft pruning to reduce irrelevant entities and uses graph-aware prompt tuning to aid LLM interpretation of topological structures. StructRAG~\cite{li2024structrag} dynamically selects optimal graph schemas for specific tasks, and KAG ~\cite{liang2024kag} constructs domain expert knowledge using conceptual semantic reasoning and human-annotated schemas, reducing noise from OpenIE systems. Despite these advances, GraphRAG paradoxically underperforms vanilla RAG on real-world tasks, with significant accuracy drops on simple queries and prohibitive latency due to its rigid application across all query types ~\cite{han2025rag,zhou2025depth}.

\begin{figure*}
    \centering
    \includegraphics[width=\linewidth]{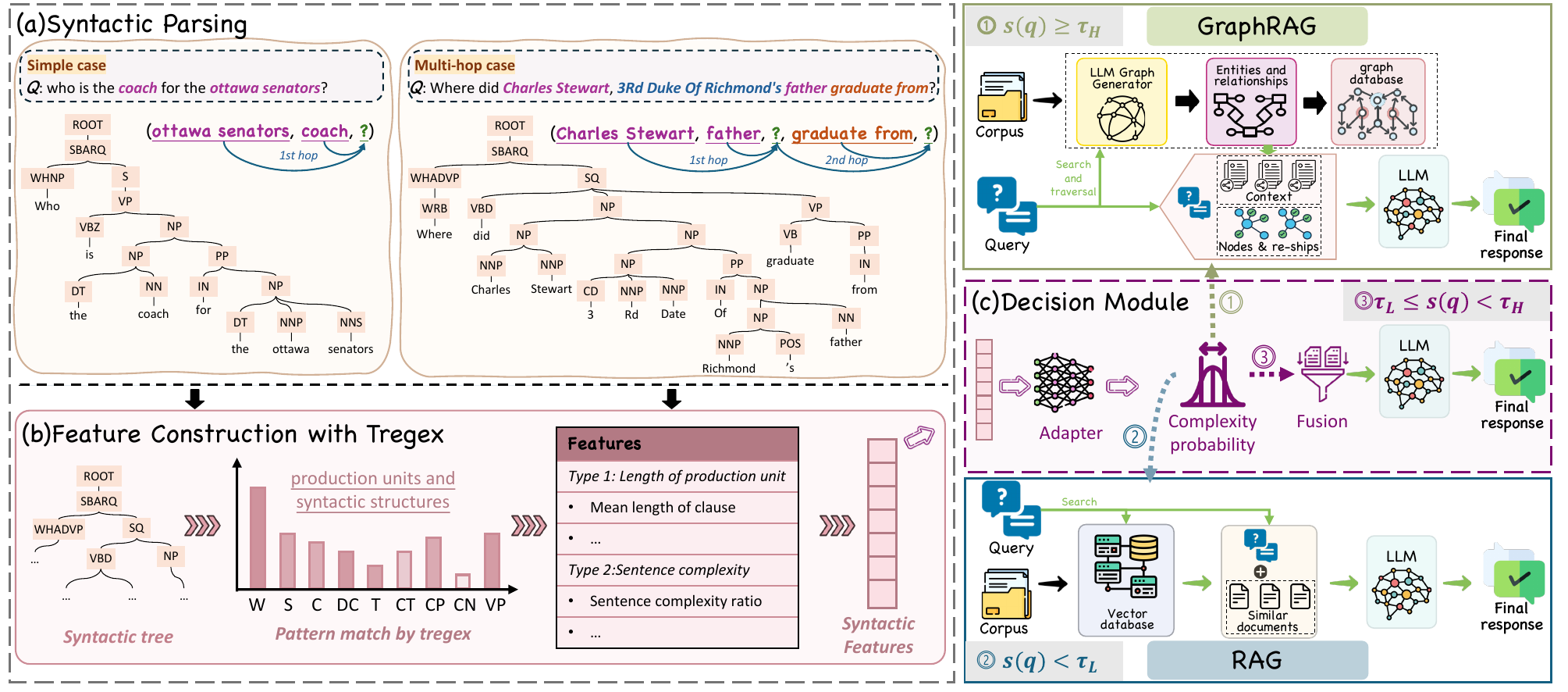}
    \caption{\textbf{The overview of our EA-GraphRAG framework.}
In the (c) Decision Module stage, an MLP adapter produces a \emph{scalar} complexity score \(s(q)\in(0,1)\). Two thresholds route the query: \(s(q)<\tau_L\Rightarrow\) Dense retrieval; \(s(q)\ge\tau_H\Rightarrow\) Graph-based retrieval; \(\tau_L\le s(q)<\tau_H\Rightarrow\) Fusion, where the documents retrieved by dense retrieval and graph-based retrieval are merged by weighted Reciprocal Rank Fusion (wRRF).}

    \label{fig:placeholder}
\end{figure*}

\subsection{Query Complexity Measurement}
Quantifying query complexity is crucial for adaptive retrieval strategy selection. Existing approaches~\cite{han2025rag,jeong2024adaptive} typically define query complexity from a semantic perspective, either by training dedicated language models or by leveraging large language models to classify queries into different complexity levels. Although effective, these methods require extensive annotated data and incur high computational costs, making them inefficient for large-scale systems. 

In contrast, the field of linguistics provides a complementary perspective: syntactic complexity has long been studied as a measurable property of sentences, reflecting the structural difficulty of language. For instance, \citet{lu2010automatic} proposed 14 syntactic complexity measures based on clauses, T-units, and complex nominals, which have been validated for distinguishing language proficiency. Inspired by this, we posit that syntactic features offer an efficient and low-cost alternative for modeling query complexity. Therefore, we train a lightweight classifier using syntactic features to predict query complexity, providing a scalable foundation for adaptive routing between retrieval strategies.

\paragraph{Limitations of Existing Approaches}Current methods suffer from two critical limitations: (1) static application of RAG or GraphRAG regardless of query complexity, leading to inefficiencies on simple tasks and underperformance on complex ones; (2) lack of integration between syntactic complexity metrics and adaptive strategies selection mechanisms. This study addresses these gaps with EA-GraphRAG, which dynamically routes queries based on syntax-aware complexity analysis, bridging RAG and GraphRAG to optimize both accuracy and latency.

\section{EA-GraphRAG}
\label{sec:framework}

We consider the retrieval-augmented question answering with a complexity-aware tri-routing architecture. Given a question \(q\) in natural language, the system retrieves relevant passages from a corpus \(\mathcal{C}=\{c_1,c_2,\ldots,c_N\}\), where \(N\) is the total number of passages and each \(c_i\) is a text passage that can be independently indexed and retrieved. Retrieved passages serve as evidence for an LLM generator \(\mathcal{G}\) to produce answers.

Our framework provides three retrieval paths:
(i) a \emph{dense} retriever \(\mathcal{R}_{\mathrm{rag}}\) that returns a ranked list \(\pi_{\mathrm{rag}}(q)\subset\mathcal{C}\) of passages ordered by semantic similarity,
(ii) a \emph{graph-based} retriever \(\mathcal{R}_{\mathrm{gr}}\) operating on a pre-built graph \(G=(V,E)\) over \(\mathcal{C}\), which returns \(\pi_{\mathrm{gr}}(q)\subset\mathcal{C}\) by combining structure-aware signals with semantic similarity, and
(iii) a \emph{complexity-aware fusion} operator \(\mathcal{F}_{\mathrm{fus}}\) that merges two ranked lists via weighted Reciprocal Rank Fusion (wRRF), producing \(\pi_{\mathrm{fus}}(q)=\mathcal{F}_{\mathrm{fus}}\big(\pi_{\mathrm{rag}}(q),\pi_{\mathrm{gr}}(q),s(q)\big)\), where fusion weights are determined by the complexity score \(s(q)\).

Routing among these paths is controlled by a decision module. A featurizer \(\phi(q)\in\mathbb{R}^d\) extracts syntactic-complexity features from the query, where \(d\) is the feature dimensionality. An MLP adapter \(g_\theta\) with feature attention and residual connections maps these features to a complexity score \(s(q)\in(0,1)\), where \(\theta\) denotes the model parameters. If \(s(q) \ge \tau_H\), the graph-based retrieval path is activated. If \(s(q) \le \tau_L\), the dense retrieval path is chosen. Otherwise, the complexity-aware fusion path is selected, where contributions from dense and graph retrieval are weighted by \(s(q)\) and \(1-s(q)\), respectively.

Finally, the answer is generated by a fixed LLM \(\mathcal{G}\) given \(q\) and the packed context. This formulation separates evidence units, retrieval operators, and a learnable complexity-aware router that selects the most appropriate path per query, enabling multi-hop reasoning when needed with smooth interpolation between methods in the fusion regime.

\begin{algorithm*}[t]
\caption{The inference pipeline of our EA-GraphRAG framework}
\label{alg:ea_graphrag_concise}

\begin{tabularx}{\textwidth}{@{}p{0.485\textwidth} p{0.485\textwidth}@{}}

\begin{algorithmic}[1]
\Require query $q$; thresholds $(\tau_L,\tau_H)$; graph $G=(V,E)$ where $V=N\cup\mathcal{C}$; fact set $\mathcal{F}$; encoder $M$; damping $\alpha$; retrieved chunks $K$; RRF constant $k$; generator $\mathcal{G}$
\Ensure answer $\hat{y}$

\State Extract features: $\mathbf{x}\gets \phi(q)$
\State Compute complexity score: $s(q) \gets \sigma(\ell_\theta(\phi(q))) \in (0,1)$

\If{$s(q)\ge\tau_H$} \Comment{Graph-based retrieval}
  \State Compute fact similarities: $s_i\gets \langle M(q), M(f_i)\rangle$ for $f_i\in\mathcal{F}$
  \State Select and rerank top-$k$ facts: $\mathcal{F}_{\text{top}}\gets \mathrm{Rerank}(\mathrm{TopK}(\tilde{\mathbf{s}}, k), q)$
  \State Compute entity weights from facts: $w(n)\gets \frac{ \tilde{s}_f}{|\{c:(n,c)\in E\}|}$ for $n\in N$
  \State Select seed entities: $Q\gets \mathrm{TopK}(\{w(n):n\in N\}, k')$
  \State Initialize PPR: $\mathbf{r}_0[v]\gets w(v)$ if $v\in Q$, else $0$; $\mathbf{r}\gets \mathrm{PPR}(G, \mathbf{r}_0, \alpha)$
  \State Extract passage scores and rank: $L\gets \mathrm{TopK}(\mathbf{r}[\mathcal{C}], K)$

\algstore{ea_split}
\end{algorithmic}
&
\begin{algorithmic}[1]
\algrestore{ea_split}

\ElsIf{$s(q)\le\tau_L$} \Comment{Dense retrieval}
  \State Encode query: $\mathbf{q}\gets M(q)$
  \State Compute similarities and rank: $L\gets \mathrm{TopK}(\{\langle\mathbf{q}, M(c)\rangle:c\in\mathcal{C}\}, K)$

\Else \Comment{Complexity-aware fusion}
  \State Dense retrieval: $\mathbf{q}\gets M(q)$; $L_r\gets \mathrm{TopK}(\{\langle\mathbf{q}, M(c)\rangle:c\in\mathcal{C}\}, K)$
  \State Graph retrieval: compute $\mathcal{F}_{\text{top}}$, $Q$, $\mathbf{r}$ as in graph path; $L_g\gets \mathrm{TopK}(\mathbf{r}[\mathcal{C}], K)$
  \State Compute fusion weights: $w_{\mathrm{gr}}\gets s(q)$; $w_{\mathrm{rag}}\gets 1-s(q)$
  \State Weighted RRF: for $c\in L_r\cup L_g$, $\mathrm{RRF}(c)\gets w_{\mathrm{rag}}\cdot\frac{1}{k+r_r(c)}+w_{\mathrm{gr}}\cdot\frac{1}{k+r_g(c)}$
  \State Fused ranking: $L\gets \mathrm{TopK}(\{\mathrm{RRF}(c):c\in L_r\cup L_g\}, K)$
\EndIf

\State Pack context and generate: $\hat{y}\gets \mathcal{G}(q, \mathrm{Pack}(L))$
\State \Return $\hat{y}$

\end{algorithmic}
\\
\end{tabularx}
\end{algorithm*}

\subsection{Offline Indexing}
\label{sec:offline-indexing}

Following~\cite{gutiérrez2025hipporag2}, we index the corpus \(\mathcal{C}\) as a heterogeneous graph \(G = (V, E)\). The node set \(V = N \cup \mathcal{C}\), where \(N\) contains noun phrases and entities extracted from \(\mathcal{C}\) via a two-stage LLM prompting strategy. The edge set \(E\) includes three types: relation edges from OpenIE triples \((h, r, t)\) within \(N\), occurrence edges linking each entity \(n_i \in N\) to its source passage \(c_j \in \mathcal{C}\), and synonymy edges between nodes \(n, v \in N\) when \(\cos(M(n), M(v)) \ge \tau\). This structure enables efficient multi-hop retrieval while preserving passage-level context. Hyperparameters follow default settings in~\cite{gutiérrez2025hipporag2}.

\subsection{Query Complexity Measurement}
\label{sec:qcm_}

To quantify query complexity for routing decisions, we design a featurizer that captures syntactic, structural, semantic, and lexical properties. Our approach is based on the principle that query complexity manifests through multiple dimensions: structural depth through nested clauses, dependency relationships via long-distance connections, semantic richness measured by entity density, and lexical diversity.

\subsubsection{Syntactic Structure Features}
We extract features from constituency parse trees using Stanza~\cite{qi2020stanza} that capture hierarchical structure. We identify nine fundamental units using Tregex pattern matching~\cite{levy2006tregex}: words W, sentences S, clauses C, dependent clauses DC, T-units T, complex T-units CT, coordinate phrases CP, complex nominals CN, and verb phrases VP. From these counts, we compute ratio-based features including mean length measures MLS=W/S and MLT=W/T, sentence complexity C/S, subordination ratios C/T, CT/T, DC/C, and DC/T, coordination ratios CP/C, CP/T, and T/S, and phrasal sophistication CN/C, CN/T, and VP/T.

\subsubsection{Dependency-based Structural Features}
From dependency parse trees using SpaCy, we extract features capturing syntactic relationships: maximum and average dependency distances, counts of long-range dependencies with distance greater than 5, relation type counts including subject-verb, object-verb, modifier, coordination, and subordination relations, and tree imbalance metrics.

\subsubsection{Semantic and Lexical Features}
We extract named entity counts and types including person, organization, location, and date, entity density, semantic role indicators such as agents, patients, temporal, and locative roles, lexical diversity measured by unique token ratio, content-to-function word ratios, information density metrics, question-type indicators, and complexity markers including coordination, subordination, negation, and passive voice.

\subsubsection{Tree Structure and Interaction Features}
From the dependency tree, we extract global properties: maximum depth, maximum width, leaf-to-non-leaf ratios, branching factors, and depth-to-width ratios. We also compute interaction terms combining multiple dimensions, such as tokens per clause, entities per token, depth per token, and connectors per clause.

\subsubsection{Feature Construction and Selection}
The featurizer \(\phi(q)\) assembles all feature categories into a \(d'\)-dimensional vector, where \(d'\) is the raw feature dimensionality before selection and typically exceeds 80. We apply mutual information-based feature selection to identify the top-\(k\) most discriminative features for predicting whether GraphRAG outperforms RAG, where \(k\) is the number of selected features and typically equals 85. The final feature vector \(\mathbf{x} \in \mathbb{R}^{d}\) has dimensionality \(d=k\) and is standardized using z-score normalization: \(\tilde{\mathbf{x}} = (\mathbf{x}-\boldsymbol{\mu}) \oslash \boldsymbol{\sigma}\), where \(\boldsymbol{\mu}\in\mathbb{R}^{d}\) and \(\boldsymbol{\sigma}\in\mathbb{R}^{d}\) are the mean and standard deviation vectors estimated on the training data, and \(\oslash\) denotes element-wise division.


\subsection{MLP Adapter}
\label{sec:mlp-adapter}

Given the \(d\)-dimensional feature vector \(\mathbf{x}=\phi(q)\), the adapter produces a complexity score \(s(q)\in(0,1)\) that controls tri-routing. We implement \(s(q)\) with a lightweight MLP trained as a binary classifier predicting whether GraphRAG will outperform RAG on a query.

\paragraph{Architecture.}
Our MLP incorporates feature attention and residual connections. The model first applies feature-level attention to the input, then passes through hidden layers with dimensions 256, 128, and 64, with residual connections where dimensions match. The final hidden representation is processed through an output attention layer before computing the logit \(\ell_\theta(\mathbf{x})\), where \(\ell_\theta\) denotes the MLP function parameterized by \(\theta\). The complexity score is:
\begin{equation}
s(q) = \sigma(\ell_\theta(\phi(q))) \in (0,1),
\label{eq:mlp-score}
\end{equation}
where \(\sigma(t)=1/(1+e^{-t})\) is the sigmoid function that maps the logit to a probability in \((0,1)\).

\paragraph{Training.}
We form a disagreement training set \(\mathcal{T}=\{i: z^{\mathrm{rag}}_i \oplus z^{\mathrm{gr}}_i = 1\}\), where \(z^{\mathrm{rag}}_i, z^{\mathrm{gr}}_i\in\{0,1\}\) are binary indicators of whether RAG and GraphRAG answer sample \(i\) correctly as measured by automatic QA metrics, and \(\oplus\) denotes exclusive OR. Each sample \(i\in\mathcal{T}\) is labeled \(y_i=1\) if GraphRAG is correct and RAG is incorrect, and \(y_i=0\) if RAG is correct and GraphRAG is incorrect. Samples where both methods agree are excluded from training. We minimize BCE-with-logits loss with label smoothing using AdamW optimizer with cosine annealing learning rate schedule.

\paragraph{Routing.}
At inference, \(s(q)\) is mapped to a path via two thresholds \(\tau_L<\tau_H\) tuned on a validation set:
\begin{equation}
\mathrm{route}(q)=
\begin{cases}
\mathrm{GraphRAG}, & s(q) \ge \tau_H,\\[2pt]
\mathrm{RAG},      & s(q) \le \tau_L,\\[2pt]
\mathrm{Fusion},   & \tau_L < s(q) < \tau_H,
\end{cases}
\label{eq:double-threshold}
\end{equation}
where \(\tau_L\) is the low threshold and \(\tau_H\) is the high threshold. In the fusion region, we use complexity-aware RRF where weights are \(w_{\mathrm{GraphRAG}} = s(q)\) and \(w_{\mathrm{RAG}} = 1 - s(q)\), with \(w_{\mathrm{GraphRAG}}\) and \(w_{\mathrm{RAG}}\) denoting the weights for graph and dense retrieval contributions, respectively. Thresholds are selected to maximize a composite objective of validation accuracy and expected latency.


\subsection{Online Retrieval}
\label{sec:retrieval}

\subsubsection{Dense Retrieval (RAG path)}
\label{sec:dense}
We encode the query and documents using encoder \(M\), which maps text to dense vector embeddings. The similarity between query \(q\) and passage \(c\) is computed as the inner product of their embeddings: \(\langle M(q), M(c)\rangle\), where \(\langle\cdot,\cdot\rangle\) denotes the dot product. We retrieve the top-\(K\) passages with highest similarity scores:
\begin{equation}
\pi_{\mathrm{rag}}(q) = \mathrm{TopK}_{c\in\mathcal{C}}\ \langle M(q), M(c)\rangle,
\label{eq:raglist}
\end{equation}
where \(K\) is the number of passages to retrieve, and \(\mathrm{TopK}\) returns the \(K\) passages with highest scores.

\subsubsection{Graph-Based Retrieval (GraphRAG path)}
\label{sec:graph}
We perform retrieval over the heterogeneous graph \(G = (V, E)\) constructed during offline indexing, where \(V = N \cup \mathcal{C}\) with \(N\) denoting the set of noun phrases and entities, and \(\mathcal{C}\) the corpus passages. The edge set \(E\) comprises relation edges from OpenIE triples, occurrence edges linking entities to their source passages, and synonymy edges connecting semantically similar entities.

Given a query \(q\), we first compute similarity scores between the query and all facts (triples) extracted from the corpus:
\begin{equation}
s_i = \langle M(q), M(f_i)\rangle, \quad \forall f_i \in \mathcal{F},
\label{eq:fact-sim}
\end{equation}
where \(\mathcal{F}\) denotes the set of all facts, \(M\) is the encoder for fact-query matching, \(f_i\) represents the \(i\)-th fact. The similarity scores are normalized to obtain \(\tilde{\mathbf{s}} = [\tilde{s}_1, \tilde{s}_2, \ldots, \tilde{s}_{|\mathcal{F}|}]^\top\), where \(\tilde{s}_i\) is the normalized similarity score for fact \(f_i\).

We then select the top-\(k\) facts based on their similarity scores and apply an LLM-based reranker to refine the selection:
\begin{equation}
\mathcal{F}_{\text{top}} = \mathrm{Rerank}\big(\mathrm{TopK}(\tilde{\mathbf{s}}, k), q\big),
\label{eq:rerank}
\end{equation}
where \(k\) is a hyperparameter controlling the number of candidate facts, \(\mathrm{TopK}(\tilde{\mathbf{s}}, k)\) returns the top-\(k\) facts with highest normalized similarity scores, and \(\mathrm{Rerank}\) uses an LLM to rerank these candidates based on their relevance to query \(q\).

We then assign weights to entity nodes based on the selected facts. For each fact \(f = (h, r, t) \in \mathcal{F}_{\text{top}}\), we extract its head entity \(h\) and tail entity \(t\), and locate the corresponding entity nodes in the graph. For each entity node \(n \in N\) that appears in \(\mathcal{F}_{\text{top}}\), we assign a weight computed as the normalized similarity score of the fact containing the entity, divided by the number of passages that contain this entity. This normalization prevents hub entities that appear in many passages from dominating the ranking. If an entity appears in multiple facts during iteration, we update its weight using the similarity score from the most recently encountered fact. After processing all facts, we select the top-\(k'\) entity nodes with the highest weights to form the seed set \(Q \subset N\), where \(k'\) is a hyperparameter controlling the number of seed entities.

We initialize the reset probability vector \(\mathbf{r}_0\) for Personalized PageRank by assigning the computed weights to entity nodes in \(Q\), while setting all passage nodes in \(\mathcal{C}\) to zero. This initialization strategy ensures that relevance signals originate exclusively from the selected entity nodes and propagate through the graph structure via relation edges connecting related entities and synonymy edges connecting semantically similar entities, eventually reaching passage nodes through occurrence edges that link entities to their source passages.

We then perform Personalized PageRank with reset probability \(\mathbf{r}_0\):
\begin{equation}
\mathbf{r} = \mathrm{PPR}(G, \mathbf{r}_0, \alpha),
\label{eq:ppr}
\end{equation}
where \(\mathbf{r} \in \mathbb{R}^{|V|}\) is the resulting diffusion vector, \(\mathrm{PPR}\) denotes the Personalized PageRank algorithm, and \(\alpha \in (0,1)\) is the damping factor controlling the teleportation probability, which is the probability of resetting to seed nodes. The PPR algorithm propagates weights from seed entity nodes through relation edges and synonymy edges within \(N\), and then diffuses to passage nodes in \(\mathcal{C}\) via occurrence edges.

Finally, we extract relevance scores for passage nodes directly from the diffusion vector \(\mathbf{r}\) and select the top-\(K\) passages:
\begin{equation}
\pi_{\mathrm{gr}}(q) = \mathrm{TopK}\big(\mathbf{r}[\mathcal{C}], K\big),
\label{eq:grlist}
\end{equation}
where \(\mathbf{r}[\mathcal{C}]\) denotes the subvector of \(\mathbf{r}\) corresponding to passage nodes, which are the entries for nodes in \(\mathcal{C}\), and \(K\) is the number of passages to retrieve. The selected passages are then passed to the generator for answer synthesis.

\begin{table*}[ht]
\centering
\caption{QA performance (Acc. and GPT-Acc.). Best per column in \textbf{bold}; second-best \underline{underlined}.}
\setlength{\tabcolsep}{8pt}
\begin{tabular}{lccccccccccc}
\toprule
& \multicolumn{4}{c}{Simple QA} & \multicolumn{4}{c}{Multi-Hop QA} & \multicolumn{2}{c}{\multirow{2}{*}{Mix}}  \\
\cmidrule(lr){2-5} \cmidrule(lr){6-9} 
Method & \multicolumn{2}{c}{NQ} & \multicolumn{2}{c}{PopQA} & \multicolumn{2}{c}{HotpotQA} & \multicolumn{2}{c}{2Wiki} & & \\
\cmidrule(lr){2-3} \cmidrule(lr){4-5} \cmidrule(lr){6-7} \cmidrule(lr){8-9} \cmidrule(lr){10-11}
& Acc. & GPT-Acc. & Acc. & GPT-Acc. & Acc. & GPT-Acc. & Acc. & GPT-Acc. & Acc. & GPT-Acc. \\
\midrule
\rowcolor{gray!20}\multicolumn{11}{c}{\textit{\textbf{Large Language Models}}} \\
Llama-3-8B~\cite{dubey2024llama} & 28.2 & 30.8 & 9.3 & 8.1 & 16.4 & 19.9 & 24.5 & 11.6 & 19.6 & 17.6 \\
Qwen3-8B~\cite{yang2025qwen3} & 37.6 & 38.7 & 24.4 & 22.8 & 24.3 & 28.8 & 35.4 & 25.3 & 30.5 & 28.9 \\
GPT-3.5-turbo & 56.5 & 62.5 & 36.1 & 35.3 & 37.2 & 45.9 & 37.2 & 33.2 & 41.8 & 44.3 \\
GPT-4o-mini~\cite{hurst2024gpt} & 54.2 & 61.1 & 34.3 & 33.8 & 34.8 & 44.1 & 34.3 & 34.8 & 39.5 & 43.4 \\
\midrule
\rowcolor{gray!20}\multicolumn{11}{c}{\textit{\textbf{RAG Baselines}}} \\
BM25~\cite{robertson1994some} & 62.3 & 67.5 & 50.4 & 51.3 & 56.1 & 68.2 & 48.0 & 52.9 & 54.2 & 59.9 \\
Contriever~\cite{izacard2021unsupervised} & 64.9 & 68.1 & 68.8 & 68.2 & 53.6 & 65.5 & 37.3 & 46.3 & 56.1 & 62.2 \\
ColBERTv2~\cite{santhanam2022colbertv2} & \underline{68.7} & \underline{71.7} & 72.8 & 71.1 & 63.4 & 75.9 & 56.4 & 62.7 & 65.3 & 70.2 \\
\midrule
\rowcolor{gray!20}\multicolumn{11}{c}{\textit{\textbf{GraphRAG Baselines}}} \\
RAPTOR~\cite{guo2024lightrag} & 62.4 & 66.7 & 55.7 & 53.5 & 60.7 & 72.0 & 34.7 & 44.3 & 53.4 & 59.1 \\
G-retriever~\cite{he2024g} & 66.4 & 67.1 & 28.8 & 28.3 & 45.3 & 51.6 & 50.6 & 32.1 & 47.7 & 44.7 \\
LightRAG~\cite{guo2025lightrag} & 64.1 & 69.4 & 68.5 & 68.1 & 59.3 & 72.2 & 46.9 & 57.4 & 59.6 & 66.7 \\
KGP~\cite{wang2024knowledge} & 61.3 & 67.6 & 52.6 & 51.6 & 54.8 & 63.7 & 43.6 & 53.8 & 53.1 & 59.2 \\
HippoRAG~\cite{HippoRAG} & 57.2 & 61.7 & 70.6 & 70.4 & 57.6 & 70.6 & \underline{68.6} & \underline{73.8} & 63.5 & 69.1 \\
HippoRAG2~\cite{gutiérrez2025hipporag2} & 67.7 & 70.9 & \underline{73.2} & \underline{71.7} & \underline{65.5} & \underline{79.5} & 67.7 & 72.5 & \underline{68.5} & \underline{73.6} \\
\midrule
\textbf{Ours} &
\textbf{69.1} & \textbf{72.6} &
\textbf{75.1} & \textbf{73.5} &
\textbf{65.9} & \textbf{80.2} &
\textbf{76.3} & \textbf{81.5} &
\textbf{71.6} & \textbf{76.9} \\
\bottomrule
\end{tabular}
\label{tab:qa_performance}
\end{table*}

\subsubsection{Fusion for Uncertain Cases (Fusion path)}
\label{sec:fusion}
When \(\mathrm{route}(q)=\mathrm{Fusion}\), we fuse \(\pi_{\mathrm{rag}}\) and \(\pi_{\mathrm{gr}}\) using complexity-aware RRF. The weighted RRF score for document \(c\) is:
\begin{equation}
\mathrm{RRF}(c) = (1-s(q)) \cdot \frac{1}{k + r_{\mathrm{rag}}(c)} + s(q) \cdot \frac{1}{k + r_{\mathrm{gr}}(c)},
\label{eq:rrf}
\end{equation}
where \(r_{\mathrm{rag}}(c)\) and \(r_{\mathrm{gr}}(c)\) are the 1-based ranks of document \(c\) in the dense retrieval list \(\pi_{\mathrm{rag}}(q)\) and graph retrieval list \(\pi_{\mathrm{gr}}(q)\), respectively. If \(c\) does not appear in a list, its rank contribution is zero. Here \(k>0\) is the RRF smoothing constant, and \(s(q)\) is the complexity score. Higher complexity queries with \(s(q)\) closer to 1 give more weight to graph retrieval, while lower complexity queries with \(s(q)\) closer to 0 favor dense retrieval. The fused list is:
\begin{equation}
\pi_{\mathrm{fus}}(q) = \mathrm{TopK}_{c \in \pi_{\mathrm{rag}}(q) \cup \pi_{\mathrm{gr}}(q)} \ \mathrm{RRF}(c),
\label{eq:hyblist}
\end{equation}
where the union \(\pi_{\mathrm{rag}}(q) \cup \pi_{\mathrm{gr}}(q)\) contains all documents appearing in either list, and \(\mathrm{TopK}\) returns the \(K\) documents with highest RRF scores.

\paragraph{Final evidence set and answer generation.}
The evidence set \(D(q)\) is determined by the routing decision:
\begin{equation}
D(q)=
\begin{cases}
\pi_{\mathrm{rag}}(q), & \text{if }\mathrm{route}(q)=\mathrm{RAG},\\[2pt]
\pi_{\mathrm{gr}}(q),  & \text{if }\mathrm{route}(q)=\mathrm{GraphRAG},\\[2pt]
\pi_{\mathrm{fus}}(q), & \text{if }\mathrm{route}(q)=\mathrm{Fusion},
\end{cases}
\label{eq:Dq}
\end{equation}
where \(D(q) \subset \mathcal{C}\) is the set of retrieved passages. The final answer is generated by:
\begin{equation}
\hat{y} = \mathcal{G}(q, \mathrm{Pack}(D(q))),
\label{eq:gen}
\end{equation}
where \(\mathcal{G}\) is a fixed LLM generator, \(\mathrm{Pack}(D(q))\) packs the passages in \(D(q)\) into a context string with truncation to fit the model's input length budget, and \(\hat{y}\) is the generated answer.


\section{Experiments}
In this section, we conduct comprehensive experiments on four benchmark datasets to verify the effectiveness and efficiency of our framework. 
\subsection{Baselines}
We selected three categories of models as our baselines: (1) Popular large language models, including Llama-3-8B~\cite{dubey2024llama}, Qwen3-8B~\cite{yang2025qwen3}, GPT-4o-mini~\cite{hurst2024gpt}, and GPT-3.5-turbo. (2) Strong and widely used retrieval methods to form RAG baselines: BM25~\cite{robertson1994some}, Contriever~\cite{izacard2021unsupervised}, and ColBERTv2~\cite{santhanam2022colbertv2}. (3) Several popular and strong GraphRAG baselines, including RAPTOR~\cite{sarthi2024raptor}, G-retriever~\cite{he2024g}, LightRAG~\cite{guo2025lightrag}, KGP~\cite{wang2024knowledge}, HippoRAG~\cite{HippoRAG} and HippoRAG2~\cite{gutiérrez2025hipporag2}.

\begin{table}[bp]
\centering
\caption{Dataset statistics}
\small 
\begin{tabular}{lccccc} 
\toprule
 & \textbf{NQ} & \textbf{PopQA} & \textbf{2Wiki} & \textbf{HotpotQA} & \textbf{Mix} \\
\midrule
\textbf{Num of queries} & 1,000 & 1,000 & 1,000 & 1,000 & 4,000 \\
\textbf{Num of docs} & 9,633 & 8,676 & 6,119 & 9,811 & 34,239 \\
\bottomrule
\end{tabular}
\label{tab:dataset_stats}
\end{table}

\subsection{Datasets}
To simulate realistic scenarios with queries of varying complexity levels, we employ both single-hop and multi-hop QA datasets in our experiments. Single-hop QA datasets effectively evaluate a RAG system's information retrieval and question answering capabilities for factual queries, while multi-hop QA datasets assess its reasoning ability when processing complex queries requiring multi-document evidence. To emulate real-world scenarios that contain a mixture of both simple and complex questions, we constructed a mixed benchmark comprising two single-hop and two multi-hop datasets used in our paper. This experimental setup comprehensively reveals the comparative strengths and limitations of RAG and GraphRAG, reflecting their performance in realistic scenarios. Table ~\ref{tab:dataset_stats} presents the statistical characteristics of our sampled dataset.

\begin{table*}[ht]
\centering
\caption{Retrieval recall (R@3 / R@5) shown as percentages with one decimal. Best per column in \textbf{bold}; second-best \underline{underlined}.}
\setlength{\tabcolsep}{10pt}
\begin{tabular}{lccccccccccc}
\toprule
& \multicolumn{4}{c}{Simple QA} & \multicolumn{4}{c}{Multi-Hop QA} & \multicolumn{2}{c}{\multirow{2}{*}{Mix}} \\
\cmidrule(lr){2-5} \cmidrule(lr){6-9}
Method & \multicolumn{2}{c}{NQ} & \multicolumn{2}{c}{PopQA} & \multicolumn{2}{c}{HotpotQA} & \multicolumn{2}{c}{2Wiki} & & \\
\cmidrule(lr){2-3} \cmidrule(lr){4-5} \cmidrule(lr){6-7} \cmidrule(lr){8-9} \cmidrule(lr){10-11}
& R@3 & R@5 & R@3 & R@5 & R@3 & R@5 & R@3 & R@5 & R@3 & R@5 \\
\midrule
\rowcolor{gray!20}\multicolumn{11}{c}{\textit{\textbf{RAG Baselines}}} \\
BM25~\cite{robertson1994some}
& 37.2 & 51.9 & 24.8 & 29.8 & 62.6 & 71.8 & 57.3 & 61.4 & 45.5 & 53.7 \\
Contriever~\cite{izacard2021unsupervised}
& 36.2 & 50.9 & 34.3 & 42.4 & 60.6 & 68.5 & 45.1 & 50.9 & 44.0 & 50.9 \\
ColBERTv2~\cite{santhanam2022colbertv2}
& 47.4 & 65.4 & 47.0 & 48.9 & 71.5 & 77.8 & 63.9 & 67.6 & 57.4 & 64.9 \\
\midrule
\rowcolor{gray!20}\multicolumn{11}{c}{\textit{\textbf{GraphRAG Baselines}}} \\
RAPTOR~\cite{guo2024lightrag}
& 44.3 & 59.7 & 28.9 & 29.3 & 45.9 & 50.1 & 33.4 & 34.9 & 38.1 & 43.5 \\
HippoRAG~\cite{HippoRAG}
& 29.6 & 43.4 & 44.6 & \underline{52.8} & 68.9 & 76.2 & 77.6 & \underline{86.0} & 55.2 & 64.7 \\
HippoRAG2~\cite{gutiérrez2025hipporag2}
& \underline{52.3} & \underline{70.8} & \underline{48.0} & 50.9 & \textbf{82.7} & \textbf{89.6} & \underline{79.4} & 85.2 & \underline{65.6} & \underline{74.1} \\
\midrule
\textbf{Ours}
& \textbf{53.9} & \textbf{71.3} & \textbf{49.2} & \textbf{54.1} & \underline{80.5} & \underline{87.2} & \textbf{80.9} & \textbf{88.1} & \textbf{66.0} & \textbf{74.9} \\
\bottomrule
\end{tabular}
\label{tab:retrieval_recall}
\end{table*}

\subsubsection{Single-hop QA} For single-hop queries, we employ two widely adopted benchmarks: Natural Questions (NQ)~\cite{kwiatkowski2019natural} and PopQA~\cite{mallen2022not}. Following the data processing methodology of ~\cite{gutiérrez2025hipporag2}, we randomly sample 1,000 queries and corresponding passages from each benchmark, ensuring a balanced and representative subset. The NQ dataset provides real user queries with a wide range of topics, while PopQA offers open-domain queries, covering diverse factual retrieval scenarios.
\subsubsection{Multi-hop QA} For multi-hop queries, we select 2WikiMultihopQA~\cite{ho2020constructing} and HotpotQA~\cite{yang2018hotpotqa}, both established as gold standards for evaluating complex reasoning capabilities of QA systems. From each dataset, we extract 1,000 queries and corresponding passages that explicitly require multi-passage reasoning, where answers must be synthesized across at least two distinct documents.

For the mixed benchmark, we pool the above datasets, uniformly mix equal numbers of single-hop (NQ/PopQA) and multi-hop (HotpotQA/2Wiki) queries, and merge their corpus to emulate a real-world query distribution.

\subsection{Metrics}
Evaluation is conducted from both answer accuracy and retrieval performance perspectives. For answer evaluation, exact string matching metrics, while widely adopted in QA evaluation, can be overly rigid since variations in casing, tense, grammar, or paraphrasing may incorrectly penalize semantically correct answers. We employ two complementary metrics that jointly assess surface fidelity and semantic correctness:

\noindent\textbf{Contain-Match Accuracy}, which measures whether the gold answer appears as a substring within the model’s prediction, allowing minor surface-form variations while maintaining semantic precision.

\noindent\textbf{GPT-Evaluation Accuracy}, an LLM-based criterion in which the evaluator receives the question, gold answer, and prediction, and determines whether the predicted answer is semantically equivalent to the reference.

Together, these metrics offer a balanced and interpretable assessment by combining the reproducibility of string matching with the nuanced semantic understanding of LLM-based judgment.

For retrieval evaluation, we report two complementary metrics. Retrieval Time per Query (s) quantifies efficiency by measuring the average time the retriever needs to return candidate documents, reflecting end-to-end retrieval latency. Recall@k assesses quality as the proportion of queries for which at least one relevant document appears among the top-k results. Together, these metrics characterize the accuracy–efficiency trade-off and offer a more holistic view of system performance.
\subsection{Implementation Details}
Our experiments were conducted on a server equipped with six RTX-3090 GPUs. For a fair comparison, all RAG and GraphRAG baseline models were evaluated under the same experimental setup, employing the GPT-4o-mini model for both question answering and graph construction, with the number of retrieved chunks (top-k) set to 5. We split the 4000-query dataset into a training set of 1000 queries and a test set of 3000 queries. From the 1000 training queries, we selected 200 samples on which RAG and GraphRAG produced different predictions, and used these disagreement cases to train the adapter to assign complexity scores for routing queries to the appropriate retrieval path.


\subsection{Main Results}
We now present the main experimental results for question answering and retrieval, in which the QA model utilizes retrieved passages as context.

\subsubsection{QA Performance}
As shown in Table~\ref{tab:qa_performance}, the proposed EA-GraphRAG framework achieves consistent improvements on both single-hop and multi-hop QA tasks, and obtains the best overall performance on the mixed benchmark with \textbf{Acc.} 71.6 and \textbf{GPT-Acc.} 76.9. These results demonstrate that the framework can effectively adapt to questions with different levels of complexity.

Compared with strong language model baselines such as GPT-4o-mini, EA-GraphRAG achieves substantial gains across all datasets. On the mixed benchmark, it improves Acc. from 39.5 to 71.6 and GPT-Acc. from 43.4 to 76.9, indicating that external retrieval remains indispensable even for powerful generators.

When compared with dense retrieval baselines, EA-GraphRAG shows clear advantages, especially on 2Wiki, where relational reasoning across multiple passages is critical. Relative to the strongest dense baseline ColBERTv2, EA-GraphRAG improves 2Wiki Acc. from 56.4 to 76.3 and GPT-Acc. from 62.7 to 81.5. Compared with GraphRAG baselines, EA-GraphRAG attains the highest performance across all datasets. It slightly improves over HippoRAG2 on HotpotQA, and delivers large gains on 2Wiki in both Acc. and GPT-Acc., validating the effectiveness of complexity-aware routing.

Overall, EA-GraphRAG achieves state-of-the-art results in both Acc. and GPT-Acc. across single-hop, multi-hop, and mixed settings. The results confirm that dynamically selecting between dense retrieval and graph-based retrieval, with a fusion fallback for borderline cases, successfully combines their complementary strengths for more robust question answering.

\subsubsection{Retrieval Performance}
We evaluate whether our routing and fusion strategy improves retrieval recall in Table~\ref{tab:retrieval_recall}. On the mixed benchmark, EA-GraphRAG achieves the best overall recall with R@3 66.0 and R@5 74.9. This result exceeds the strongest dense baseline ColBERTv2 by 8.6 points on R@3 and 10.0 points on R@5, and improves over the strongest graph baseline HippoRAG2 by 0.4 points on R@3 and 0.8 points on R@5.

The gains are consistent across both simple and multi-hop regimes. On NQ, EA-GraphRAG achieves R@3 53.9 and R@5 71.3, outperforming HippoRAG2 by 1.6 and 0.5 points. On PopQA, it reaches R@3 49.2 and R@5 54.1, improving over ColBERTv2 by 2.2 and 5.2 points. On multi-hop QA, EA-GraphRAG obtains R@3 80.5 and R@5 87.2 on HotpotQA, and R@3 80.9 and R@5 88.1 on 2Wiki. These results indicate that complexity-aware routing and fusion surface high-recall evidence for both simple factual queries and multi-hop reasoning queries, validating our retrieval design.

We further measure per-query retrieval latency in Table~\ref{tab:retrieval-time}. The dense path is highly efficient, yielding a 96.4\% reduction versus graph retrieval. Leveraging this, our router without fusion runs at 1.14 s/query and the full system at 2.19 s/query. The overhead of the full system is mainly introduced by invoking fusion on borderline queries, reflecting a tunable accuracy--efficiency trade-off controlled by the two thresholds $\tau_L$ and $\tau_H$.

\begin{table}[h]
\centering
\caption{Average retrieval time per query (seconds).}
\small
\begin{tabularx}{\linewidth}{L C}
\toprule
\textbf{Retrieval strategy} & \textbf{Time/query (s)} \\
\midrule
dense retrieval        & 0.08 \\
graph-based retrieval  & 3.23 \\
Ours w/o fusion        & 1.14 \\
\textbf{Ours}          & \textbf{2.19} \\
\bottomrule
\end{tabularx}
\label{tab:retrieval-time}
\end{table}

\renewcommand\tabularxcolumn[1]{>{\RaggedRight\arraybackslash}p{#1}}

\newcommand{\MethodRow}[2]{%
  \textbf{#1} &
  \begin{minipage}[t]{\linewidth}
    #2
  \end{minipage}\\[2pt]
}

\begin{table*}[ht]
\centering
\caption{Cases comparing Vanilla RAG, GraphRAG and EA-GraphRAG.}
\label{tab:case_study}
\small
\setlength{\tabcolsep}{6pt}
\renewcommand{\arraystretch}{1.08}
\begin{threeparttable}
\begin{tabularx}{\textwidth}{>{\bfseries}p{0.18\textwidth} X}
\toprule
Single-hop Question & When is season 8 for game of thrones? \\
\midrule
Ground Truth & 2019 \\
\midrule

\MethodRow{GraphRAG}{
\textit{Retrieved context:}
\begin{enumerate}[leftmargin=*,itemsep=0pt,parsep=0pt,topsep=2pt,partopsep=0pt]
  \item \xmark\ ``Game of Thrones'': Game of Thrones Game of Thrones is an American fanta\ldots
  \item \xmark\ ``The Walking Dead (season 8)'': marked the first crossover between the two series\ldots
\end{enumerate}
\ldots \\
\textit{Prediction:} \xmark\ The text does not provide a specific date for Season 8 of Game of Thrones.
}

\MethodRow{EA-GraphRAG}{
\textit{Retrieved context:}
\begin{enumerate}[leftmargin=*,itemsep=0pt,parsep=0pt,topsep=2pt,partopsep=0pt]
  \item \cmark\ ``Game of Thrones (season 8)'': \ldots to premiere in \textbf{April 2019}. Ramin Djawadi\ldots
  \item \cmark\ ``Game of Thrones (season 8)'': \ldots The season is scheduled to premiere in \textbf{April 2019}\ldots
\end{enumerate}
\ldots \\
\textit{Prediction:} \cmark\ April 2019.
}

\midrule
Multi-hop Question & What is the date of death of the director of film The Organization (Film)? \\
\midrule
Ground Truth & December 12, 2012 \\
\midrule

\MethodRow{Vanilla RAG}{
\textit{Retrieved context:}
\begin{enumerate}[leftmargin=*,itemsep=0pt,parsep=0pt,topsep=2pt,partopsep=0pt]
  \item \xmark\ ``Lino Brocka'': Catalino Ortiz Brocka (April 3, 1939 – May 22, 1991) was a Filipino film director\ldots
  \item \xmark\ ``Wallace Fox'': Wallace Fox (March 9, 1895 – June 30, 1958) was an American film director\ldots
\end{enumerate}
\ldots \\
\textit{Prediction:} \xmark\ Not provided in the text.
}

\MethodRow{EA-GraphRAG}{
\textit{Retrieved context:}
\begin{enumerate}[leftmargin=*,itemsep=0pt,parsep=0pt,topsep=2pt,partopsep=0pt]
  \item \cmark\ ``The Organization (film)'': \ldots \textbf{The Organization} is a 1971 DeLuxe Color American crime thriller film starring Sidney Poitier as Virgil Tibbs and directed by \textbf{Don Medford}\ldots
  \item \cmark\ ``Don Medford'': \ldots \textbf{Donald Muller} (November 26, 1917 – \textbf{December 12, 2012})\ldots
\end{enumerate}
\ldots \\
\textit{Prediction:} \cmark\ December 12, 2012.
}

\bottomrule
\end{tabularx}
\end{threeparttable}
\vspace{-3mm}
\end{table*}

\subsection{Analysis of the Query Adapter}
\label{sec:adapter_analysis}

\begin{figure}[t]
  \centering
  \includegraphics[width=\linewidth]{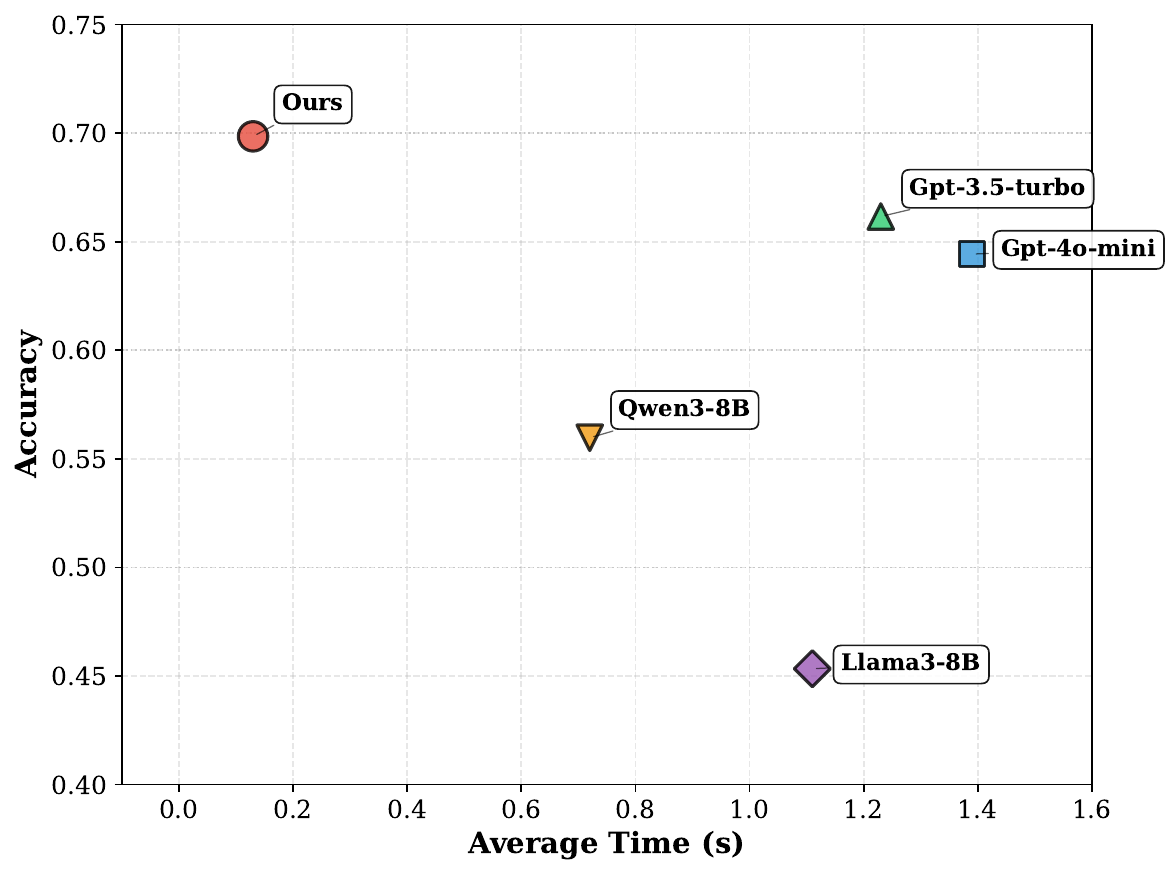}
  \caption{Accuracy and efficiency comparison of adapters.}
  \label{fig:pareto}
\end{figure}

We analyze the proposed query adapter from the perspective of end-to-end efficiency. Figure~\ref{fig:pareto} presents the accuracy--latency Pareto plot comparing our adapter-based framework with representative semantic model baselines. The adapter yields a markedly better trade-off, achieving higher accuracy while operating at substantially lower average inference time. This advantage stems from the adapter's lightweight design: it relies on inexpensive syntactic feature extraction and a small predictor to estimate query difficulty and trigger routing or fusion, avoiding the need to run an additional heavyweight semantic model for every query.

Importantly, the adapter improves not only runtime but also overall effectiveness. Semantic scorers can be accurate but often introduce non-trivial per-query overhead, which becomes prohibitive under mixed workloads and interactive settings. In contrast, our adapter enables fast complexity-aware decisions with minimal latency, allowing the system to allocate expensive graph-based retrieval only when needed and to fall back to dense retrieval for simpler cases. As a result, our framework reaches a superior operating point on the Pareto frontier, demonstrating that the proposed adapter is both practical and effective for real-world deployment.

\subsection{Ablation Study}

\begin{table}[h]
\caption{Ablation on the mixed benchmark. Starting from a generator-only baseline, progressively adding dense retrieval, graph-based retrieval, and finally the fusion module (RRF) yields steady gains; the full EA-GraphRAG achieves the best accuracy and GPT-based accuracy.}
\centering
\footnotesize
\setlength{\tabcolsep}{10pt}
\vspace{-3mm}
\begin{tabular}{lcc}
\toprule
\textbf{Ablation setting} & \textbf{Acc.} & \textbf{GPT-Acc.} \\
\midrule
GPT-4o-mini (no retrieval) & 39.5 & 43.4 \\
\quad + Dense retrieval & 65.1 & 70.1 \\
\quad + Dense \& Graph-based retrieval & 70.5 & 75.9 \\
\quad + Dense \& Graph-based retrieval + Fusion  & \textbf{71.6} & \textbf{76.9} \\
\bottomrule
\end{tabular}
\label{tab:ablation-mixed}
\vspace{-5mm}
\end{table}

To assess the contribution of each component in EA-GraphRAG, we conduct an ablation study on the mixed benchmark; results are reported in Table~\ref{tab:ablation-mixed}. The generator-only baseline achieves 39.5 Acc. and 43.4 GPT-Acc., indicating that parametric knowledge alone is insufficient for this setting. Adding dense retrieval leads to a large gain, reaching 65.1 Acc. and 70.1 GPT-Acc., which highlights the importance of grounding generation with retrieved evidence. Incorporating graph-based retrieval further improves performance to 70.5 Acc. and 75.9 GPT-Acc., showing that structural signals provide complementary benefits beyond dense retrieval. Finally, adding the fusion module yields the best results, achieving 71.6 Acc. and 76.9 GPT-Acc. This demonstrates that combining dense and graph retrieval outputs with Reciprocal Rank Fusion effectively improves robustness on queries where a single retrieval path is insufficient.

\subsection{Case Study}

Table~\ref{tab:case_study} presents two representative examples that illustrate why EA-GraphRAG outperforms using either retrieval paradigm alone. For the single-hop factoid ``\emph{When is season 8 for game of thrones?},'' the graph-based retriever HippoRAG places excessive emphasis on entity matches and link structure. As a result, it retrieves passages that contain relevant entities but do not include the local sentence-level evidence specifying the release date, which introduces off-target context and leads to an incorrect answer. Dense retrieval, by prioritizing sentence- and passage-level semantic similarity, directly surfaces passages stating that the season ``premiere[s] in April 2019,'' which is sufficient to answer the question. 

For the multi-hop query ``\emph{What is the date of death of the director of \textit{The Organization} (Film)?},'' dense retrieval tends to return generic director biographies and fails to recover the required evidence chain from the film to its director Don Medford and then to his death date. In contrast, the graph-based path uses entity linking and diffusion over the graph to traverse these relations and retrieve evidence containing ``December 12, 2012.'' EA-GraphRAG addresses these complementary failure modes by routing simple factoids to dense retrieval and compositional multi-hop questions to graph-based retrieval, while reserving the fusion path for ambiguous cases. Together, these examples show that dynamic routing enables a single framework to consistently retrieve appropriate evidence across both simple and complex queries.

\section{Conclusion}
Graph retrieval-augmented generation (GraphRAG) has recently emerged as a powerful paradigm that leverages external structured graphs to enable deep retrieval and contextual comprehension. Existing GraphRAG frameworks excel at multi-hop reasoning but struggle with simple factoid queries and are far less efficient than vanilla RAG, due to their uniform reliance on graph construction and multi-step traversal that introduces additional retrieval overhead and can inject irrelevant graph-induced noise for queries that only require localized evidence. To remedy this, we proposed \textbf{EA-GraphRAG}, which performs syntactic complexity analysis of the query and \emph{routes} simple queries to dense (RAG) retrieval while sending complex ones to graph-based retrieval, with an optional fusion fallback for borderline cases. This design preserves semantic precision on easy queries and leverages graph structure for genuinely compositional ones, enabling a single system to handle mixed difficulty levels effectively. Comprehensive experiments demonstrate that EA-GraphRAG delivers consistent gains in retrieval and end-to-end QA accuracy while improving efficiency, validating the effectiveness of complexity-aware routing in real-world, heterogeneous query distributions.


\bibliographystyle{ACM-Reference-Format}
\bibliography{main}

\end{document}